\documentclass[11pt,a4paper]{article}
\usepackage[hyperref]{emnlp2018}
\usepackage{times}
\usepackage{latexsym}
\usepackage{amsmath}
\usepackage{makecell}
\usepackage{linguex}
\usepackage{booktabs}
\usepackage{subcaption}

\usepackage{graphicx}

\usepackage{url}

\aclfinalcopy 
\def\aclpaperid{657} 

\title{A Neural Model of Adaptation in Reading}

\author{Marten van Schijndel\\
  Department of Cognitive Science\\
  Johns Hopkins University\\
  {\tt vansky@jhu.edu} \\\And
  Tal Linzen \\
  Department of Cognitive Science\\
  Johns Hopkins University\\
  {\tt tal.linzen@jhu.edu}\\}

\date{}

\begin{document}
\maketitle
\begin{abstract}
It has been argued that humans rapidly adapt their lexical and syntactic expectations to match the statistics of the current linguistic context. We provide further support to this claim by showing that the addition of a simple adaptation mechanism to a neural language model improves our predictions of human reading times compared to a non-adaptive model. We analyze the performance of the model on controlled materials from psycholinguistic experiments and show that it adapts not only to lexical items but also to abstract syntactic structures.
\end{abstract}

\setlength{\Exlabelwidth}{0.1em}
\setlength{\SubExleftmargin}{1.5em}
\section{Introduction}\label{sec:intro}
Reading involves the integration of noisy perceptual evidence with probabilistic expectations about the likely contents of the text. Words that are consistent with these expectations are identified more quickly \cite{ehrlich1981contextual,smithlevy13}. For the reader's expectations to be maximally effective, they should not only reflect the reader's past experience with the language \cite{hale2001probabilistic,macdonald2002reassessing}, but should also be \textit{adapted} to the current context. Optimal adaptation would reflect properties of the text being read, such as genre, topic and writer identity, as well as the general tendency for recently used words and syntactic structures to be reused with higher probability \cite{bock1986syntactic,church2000empirical,dubey2006integrating}.

Several studies have suggested that readers do in fact adapt their lexical and syntactic predictions to the current context \cite{ottenvanberkum08,fineetal13,finejaeger16}.\footnote{Recently, \newcite{stack2018failure} questioned the robustness of the results of \newcite{fineetal13}.} For example, \citeauthor{finejaeger16} investigated the processing of ``garden path'' sentences such as \ref{ex:experienced_ambig}:

\ex.The experienced soldiers warned about the dangers conducted the midnight raid.\label{ex:experienced_ambig}

The word \emph{warned} in \ref{ex:experienced_ambig} is initially ambiguous between a main verb interpretation (the soldiers were doing the warning) and a reduced relative clause interpretation (the soldiers were being warned).
When the word \emph{conducted} is reached, this ambiguity is resolved in favor of the reduced relative parse. Reduced relatives are infrequent constructions. This makes the disambiguating word \emph{conducted} unexpected, causing it to be read more slowly than it would be in a context such as \ref{ex:experienced_unambig}, in which the words \textit{who were} indicate early on that only the relative clause parse is possible:

\ex.The experienced soldiers who were warned about the dangers conducted the midnight raid.\label{ex:experienced_unambig}

\citeauthor{finejaeger16} included a large proportion of reduced relatives in their experiment. As the experiment progressed, the cost of disambiguation in favor of the reduced relative interpretation decreased, suggesting that readers had come to expect a construction that is normally infrequent.

Human syntactic expectations have been successfully modeled with syntax-based language models \cite{hale2001probabilistic,levy08,roarketal09}. Recently, language models (LMs) based on recurrent neural networks (RNNs) have been shown to make adequate syntactic predictions \cite{linzen2016assessing,gulordavaetal18}, and to make comparable reading time predictions to syntax-based LMs \cite{vanschijndellinzen18}. In this paper, we propose a simple way to continuously adapt a neural LM, and test the method's psycholinguistic plausibility.
We show that LM adaptation significantly improves our ability to predict human reading times using the LM. Follow-up experiments with controlled materials show that the LM adapts not only to specific vocabulary items but also to abstract syntactic constructions, as humans do.

\section{Method}
We use a simple method to adapt our LM: at the end of each new test sentence, we update the parameters of the LM based on its cross-entropy loss when predicting that sentence; the new weights are then used to predict the next test sentence.%
\footnote{Our code is publicly available at: \url{https://github.com/vansky/neural-complexity.git}}
Our baseline LM is a long short-term memory \citep[LSTM;][]{hochreiterschmidhuber97} language model trained on 90 million words of English Wikipedia by \newcite{gulordavaetal18} (see Supplementary Materials for details). For adaptation, we keep the learning rate of 20 used by \citeauthor{gulordavaetal18}\ (the gradient is multiplied by this learning rate during weight updates). We examine the effect of this parameter in Section~\ref{sec:dative}.

We tested the model on the Natural Stories Corpus \cite{futrelletal18}, which has 10 narratives with self-paced reading times from 181 native English speakers.
There are two narrative genres in the corpus: fairy tales (seven texts) and documentary accounts (three texts).

\section{Linguistic accuracy}
We first measured how well the adaptive model predicted upcoming words. We report the model's perplexity, a quantity which is lower when the LM assigns higher probabilities to the words that in fact occurred. We adapted the model to the first $k$ sentences of each text, then tested it on sentence $k+1$, for all $k$. 
Adaptation dramatically improved test perplexity compared to the non-adaptive version of the model (86.99 vs.\ 141.49).

We next adapted the model to each genre separately.
If the model adapts to stylistic or syntactic patterns, we might expect adaptation to be more helpful in the fairy tale than the documentary genre: the Wikipedia corpus that the LM was originally trained on is likely to be more similar in style to the documentary genre. Consistent with this hypothesis, the documentary texts benefited less from adaptation (99.33 to 73.20) than the fairy tales (160.05 to 86.47), though the fact that both saw improvement from adaptation suggests that text-specific adaptation is beneficial even if the genre is similar to the training genre.

Each genre consists of multiple texts. Does adaptation to a particular text lead to catastrophic forgetting \cite{mccloskeycohen89}, such that the LM overfits to the text and forgets its more general knowledge acquired from the Wikipedia training corpus? This was not the case; in fact, adapting to the entirety of each genre without reverting to the baseline model after each text led to a very slightly \textit{better} perplexity (fairytales: 86.47, documentaries: 73.20) compared with a setting in which the LM was reverted after each text (fairytales: 86.61, documentaries: 73.63).

\begin{table}[t]
    \begin{tabular}{lrrr}
      & $\hat\beta$ & $\hat\sigma$ & t\\
      \toprule
      \multicolumn{3}{l}{\textsc{Without adaptive surprisal:}}\\
  {Sentence position} & 0.55 & 0.53 & 1.03\\
  {Word length} & 7.29 & 1.00 & 7.26\\
  {Non-adaptive surprisal} & 6.64 & 0.68 & 9.79 \\\midrule
      \multicolumn{3}{l}{\textsc{With adaptive surprisal:}}\\
  {Sentence position} & 0.29 & 0.53 & 0.55\\
  {Word length} & 6.42 & 1.00 & 6.40\\
  {Non-adaptive surprisal} & -0.89 & 0.68 & -1.31 \\
  {Adaptive surprisal} & 8.45 & 0.63 & 13.42 \\\bottomrule
\end{tabular}

\caption{Fixed effect regression coefficients from fitting self-paced reading times. The top model lacks fixed and random effects of adaptive surprisal. In general, a coefficient is significant when $|t| > 2$.}
\label{tab:noadapt}
\end{table}

\section{Modeling human expectations}
We next tested whether our adaptive LM matches human expectations better than a non-adaptive model.
Since each reader saw the texts in a different order, we adapted the LM to each text separately: after each story, we reverted to the initial Wikipedia-trained LM and restarted adaptation on the next text.
If anything, this likely resulted in a conservative estimate of the benefit of adaptation compared to a model that adapts continuously across multiple stories from the same genre, as humans might do.%
\footnote{We do not distinguish between \emph{priming} and \emph{adaptation} in this paper. While it may be tempting to think of the LSTM memory cell as a model of priming and of the weight updates as a model of adaptation, \citet{bockgriffin00} provide evidence that priming cannot simply be a function of residual activation and that priming can be driven by longer-term learning (see \citet{tooleytraxler10} for more discussion on priming vs.\ adaptation).}

We used surprisal as a linking function between the LM's predictions and human reading times \cite{hale2001probabilistic,smithlevy13}. Surprisal quantifies how unpredictable each word ($w_i$) is given the preceding words:
\begin{equation}
\text{surprisal}(w_i) = -\text{log P}(w_i\mid w_1...w_{i-1})
\end{equation}
We fit the self-paced reading times in the Natural Stories Corpus with linear mixed effects models (LMEMs), a generalization of linear regression (see Supplementary Materials for details). 

In line with previous work, non-adaptive surprisal was a significant predictor of reading times ($p < 0.001$) when the model only included other baseline factors (Table~\ref{tab:noadapt}, Top). Adaptive surprisal was a significant predictor of reading times ($p < 0.001$) over non-adaptive surprisal and all baseline factors (Table~\ref{tab:noadapt}, Bottom). Crucially, non-adaptive surprisal was no longer a significant predictor of reading times once adaptive surprisal was included. This indicates that the predictions of the adaptive model subsume the predictions of the non-adaptive one.

  \begin{figure}[t]
    \centering
    \includegraphics[width=0.475\textwidth]{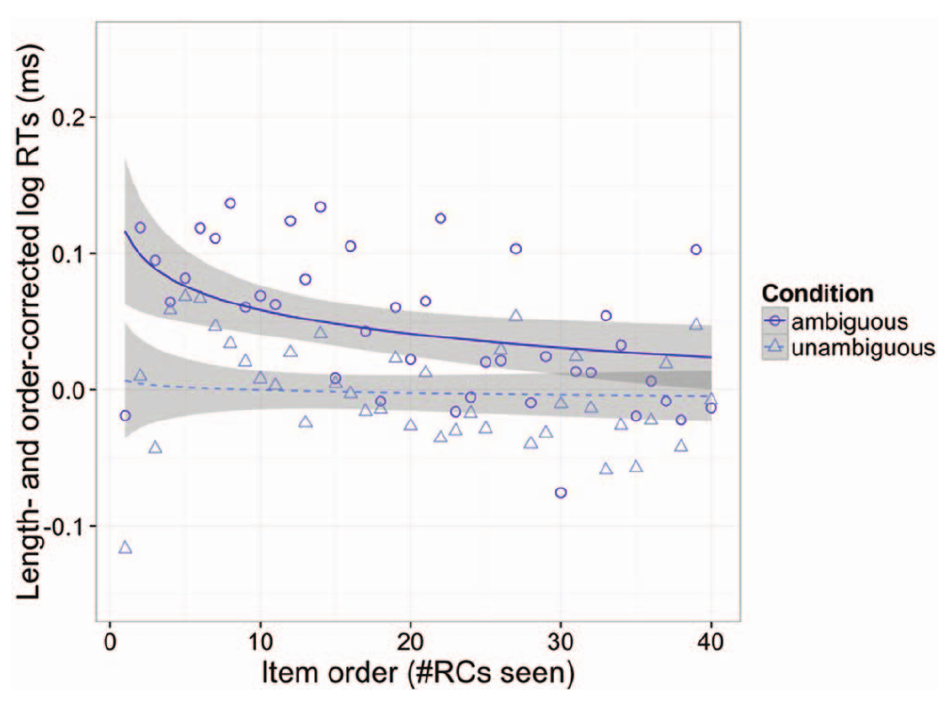}
    \caption{Mean length- and order-corrected reading times over the disambiguating region of the critical items in \citet{finejaeger16}. Figure adopted from that paper.}\label{fig:origfineresults}
  \end{figure}

  \section{Does the model adapt to syntax?}

We have shown that LM adaptation improves our ability to model human expectations as reflected in a self-paced reading time corpus. How much of this improvement is due to adaptation of the model's syntactic representations \cite{bacchiani2006map,dubey2006integrating} and how much is simply due to the model assigning a higher probability to words that have recently occurred \cite{kuhndemori90,church2000empirical}? We address this question using two syntactic phenomena: reduced relative clauses and the dative alternation.

\subsection{Reduced relative clauses}

  \begin{figure}[t]
    \centering
    \includegraphics[width=0.475\textwidth]{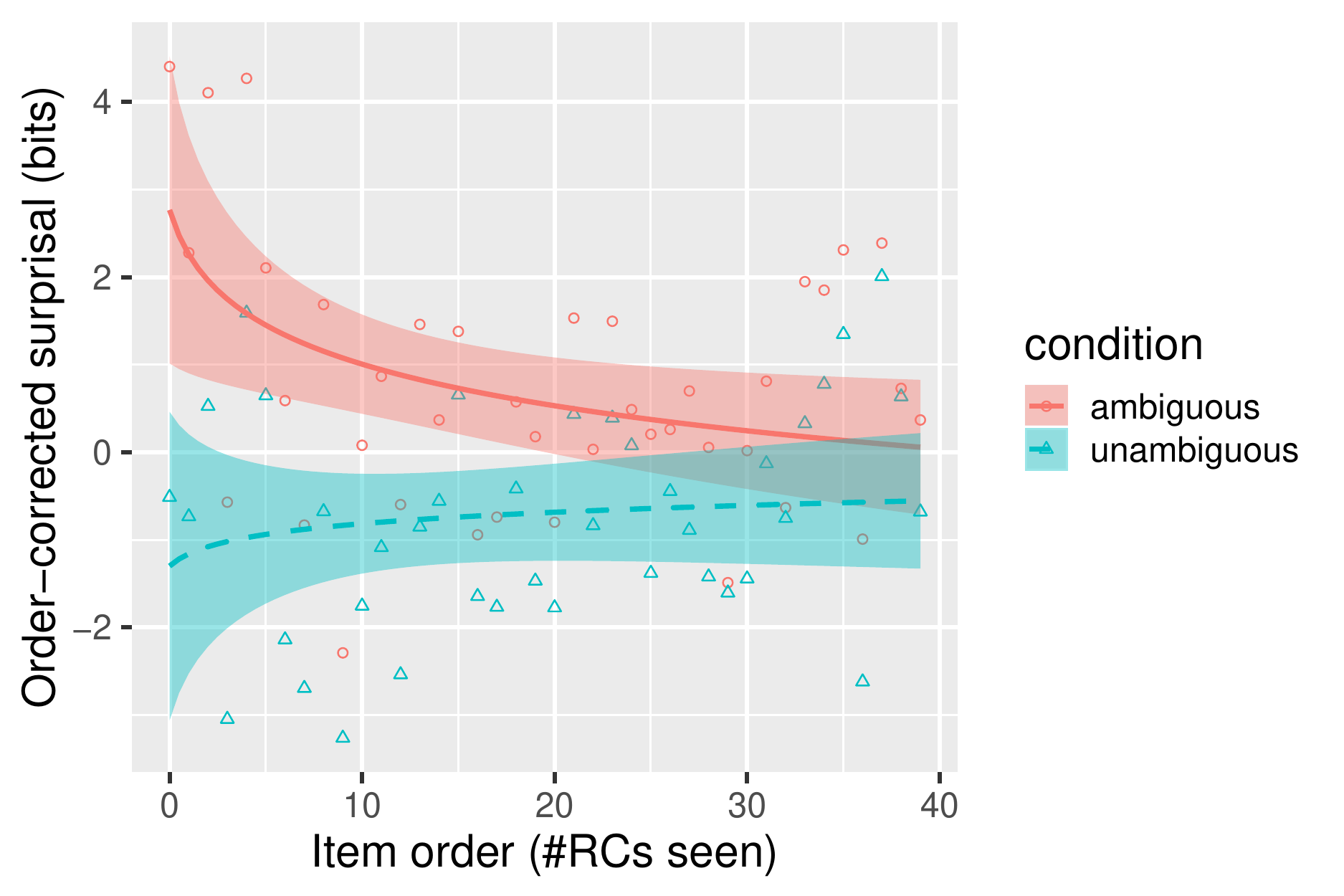}
    \caption{Mean order-corrected model surprisal over the disambiguating region of the critical items in \citet{finejaeger16}.}\label{fig:fineadapt}
  \end{figure}

We adapted the model independently to random orderings of the critical and filler stimuli used in Experiment 3 of \citet{finejaeger16};\footnote{See details in the Supplementary Materials.} this experiment (described in the Introduction) contained a much higher proportion of reduced relative clauses than their general distribution in English.
We used surprisal as our proxy for reading times. Following \citeauthor{finejaeger16}, we took the mean surprisal over three words in each ambiguous sentence: the disambiguating word and the following two words (e.g., \emph{conducted the midnight} in example~\ref{ex:experienced_ambig}). To estimate the magnitude of the syntactic disambiguation penalty while also controlling for lexical content, we subtracted this quantity from the mean surprisal over the exact same words in the paired unambiguous sentence~\ref{ex:experienced_unambig}. Linear regression showed that the disambiguation penalty decreased as the model was exposed to more critical items (item order coefficient: $\hat\beta=-0.0804$, $p<0.001$), indicating that the LM was adapting to reduced relatives, a syntactic construction without any lexical content.

In order to compare our findings more directly with the results given by \citet{finejaeger16} (shown in Figure~\ref{fig:origfineresults}), we mimicked their method of plotting reading times.
First, we fit a linear model of the mean surprisal of each disambiguating region with the number of trials the model had seen in the experiment thus far to account for a general trend of subjects speeding up over the course of the experiment.
Then, we plotted the mean residual model surprisal that was left in the disambiguating region in both the ambiguous and unambiguous conditions as the experiment progressed.
The shape of our model's adaptation to the reduced relative construction (upper curve in Figure~\ref{fig:fineadapt}) matched the human results reported by \citeauthor{finejaeger16}. Like humans, the model showed an initially large adaptation effect, followed by more gradual adaptation thereafter. Both humans and our model continued to adapt over all the items rather than just at the beginning of the experiment.
Also like humans, the model's response to unambiguous items did not change significantly over the course of the experiment (p $= 0.91$). 

  \subsection{The dative alternation}\label{sec:dative}

Dative events can be expressed using two roughly equivalent English constructions: 

\ex.\label{ex:dative}
\a.\textit{Prepositional object (PO)}:\\
      The boy threw the ball to the dog.
      \b.\textit{Double object (DO)}:\\
      The boy threw the dog the ball.
  
      Work in psycholinguistics has shown that recent experience with one of these variants increases the probability of producing that variant \cite{bock1986syntactic,kaschak2006recent} as well as the likelihood of predicting it in reading \cite{tooley2014parity}. To test whether our adaptation method can reproduce this behavior, we generated 200 pairs of dative sentences similar to~\ref{ex:dative}. We shuffled 100 DO sentences into 1000 filler sentences sampled from the Wikitext-2 training corpus \cite{wikitext-2} and adapted the model to these 1100 sentences. We then froze the weights of the adapted model and tested its predictions for two types of sentences: the PO counterparts of the DO sentences in the adaptation set, which shared the vocabulary of the adaptation set but differed in syntax; and 100 new DO sentences, which shared syntax but no content words with the adaptation set.\footnote{For additional details as well as the reverse setting (adaptation to PO), see Supplementary Materials.}

      An additional goal of this experiment was to examine the effect of learning rate on adaptation.  During adaptation the model performs a single parameter update after each sentence and does not train until convergence with gradual reduction of the learning rate as would normally be the case during LM training. Consequently, the learning rate parameter crucially determines the amount of adaptation the model can undertake after each sentence. If the learning rate is very low, adaptation will not have any effect; if it is too high, either the model will overfit after each update and will not generalize well, or the model will forget its trained representation as it overshoots the targeted minima. 
      The optimal rate may differ between lexical and syntactic adaptation. Our experiments thus far all used the same learning rate as our original model (20); here, we varied the learning rate on a logarithmic scale between 0.002 and 200.

  \begin{figure}[t]
    \includegraphics[width=.48\textwidth]{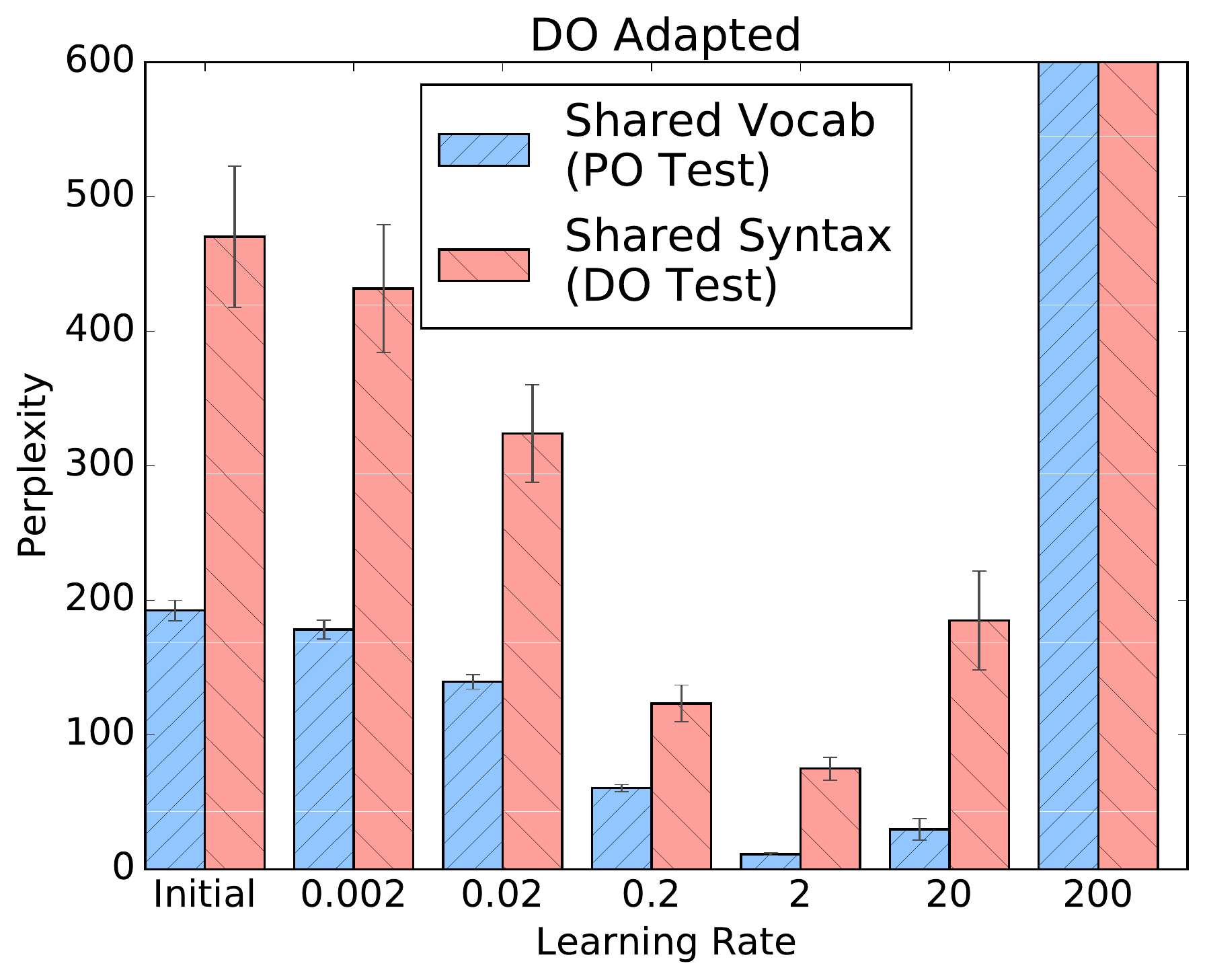}
    \caption{Learning rate influence over syntactic and lexical adaptation. The initial non-adaptive model performance is equivalent to the performance when using a learning rate of 0; the learning rate of 200 resulted in perplexity in the billions.}\label{fig:learnrate}
  \end{figure}

The results of this experiment are shown in Figure~\ref{fig:learnrate}.
The model successfully adapted to the DO construction as well as to the vocabulary of the adaptation sentences. This was the case for all of the learning rates except for 200, which resulted in enormous perplexity on both sentence types. Both lexical and syntactic adaptation were most successful when the learning rate was around~2, with perplexity reductions of 94\% for lexical adaptation and 84\% for syntactic adaptation. 

Syntactic adaption was penalized at higher learning rates more than lexical adaptation (compare learning rates of 2 and 20). 
This fragility of syntactic adaptation likely stems from the fact that the model can directly observe the relevant vocabulary but syntax is latent and must be inferred from multiple similar sentences, a generalization which is impeded by overfitting at higher learning rates.

\begin{figure}[t]
    \centering
    \includegraphics[width=.4\textwidth]{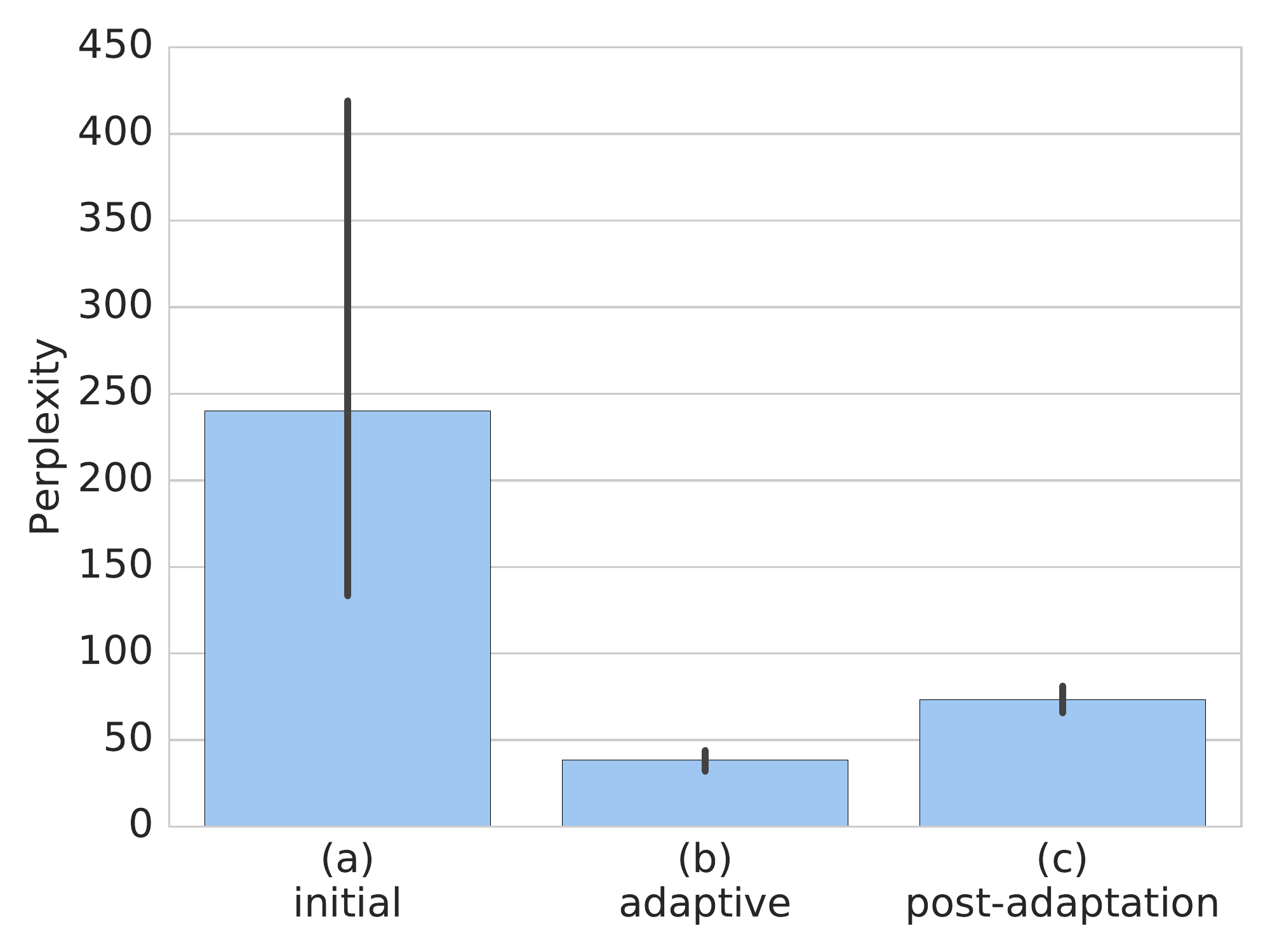}
    \caption{Perplexity on the held-out set of $G_1$ (a) before adaptation, (b) after adaptation to $G_1$, (c) after adapting to $G_1$ then adapting to $G_2$.}\label{fig:multinli}
  \end{figure}

\section{Testing for catastrophic forgetting}

Our analysis of the Natural Stories corpus did not indicate that the model suffered from catastrophic forgetting. Yet the Natural Stories corpus contained only two genres; to address the issue of catastrophic forgetting more systematically, we used the premise sentences from the MultiNLI corpus \cite{williamsetal18} --- a total of 2000 sentences for each of 10 genres.

For each genre pair $G_1$ and $G_2$ (omitting cases where $G_1 = G_2$), we first adapted the baseline Wikipedia model to 1000 sentences of $G_1$ using a learning rate of 2 (shown to be optimal in Section~\ref{sec:dative}).
We then adapted the model to 1000 sentences of $G_2$.
Finally, we froze the model's weights and tested its perplexity on the 1000 held-out sentences from $G_1$.

The results averaged across all pairs of genres are plotted in Figure~\ref{fig:multinli}.
Unsurprisingly, the model performed best on $G_1$ immediately after adapting to it (middle bar). Crucially, even after adapting to 1000 sentences of $G_2$ after its last exposure to $G_1$ (right bar), it still modeled $G_1$ much better than the non-adapted model (left bar).
These results suggest that catastrophic forgetting is not a concern even with a relatively large amount of data.

\section{Discussion}

Adaptation greatly improved an RNN LM's word prediction accuracy, in line with other work on LM adaptation \cite{knesersteinbiss93}. We showed that the adapted model was psycholinguistically plausible, in two senses. First, it improved the correlation between surprisal derived from the model and human reading times, suggesting that the model generated more human-like expectations. Second, using materials that teased apart lexical content from syntax, we showed that the model adapted both its lexical and its syntactic predictions, in line with findings from human experiments. Finally, as in other neural-network based models in psychology \cite{chang2006becoming}, our gradient-based updates naturally incorporate the error-driven nature of syntactic adaptation; while we did not demonstrate this in the current paper, we hypothesize that our model will reproduce the finding that more surprising words lead to greater adaptation \cite{jaegersnider13}.

The simplicity of our adaptation method makes it attractive for use in modeling human expectations. Since adaptive surprisal is strictly superior to non-adaptive surprisal in modeling reading times, it would be a stronger baseline in analyses that aim to demonstrate the contribution of factors other than predictability.

We used a simple neural adaptation approach, where we performed continuous gradient updates based on the prediction error on the adaptation sentences \citep[see also][]{krause2017dynamic}. An alternative approach to neural LM adaptation uses recent RNN states in conjunction with the current state to make word predictions \cite{graveetal17,merity2017pointer}; a comparison of the two methods using our paradigms may provide insight into their relative strengths and weaknesses.

Finally, we reverted to the base model after the end of each text in our experiments, forgetting any text-specific adaptation. This mimics the effect of a participant leaving an experiment that had an unusual distribution of syntactic constructions and reverting to their standard expectations. In practice, however, humans are able to generalize from prior experience when they begin adapting to a new speaker or text if it is similar in some way to their previous experiences. For example, the model of \citet{jaechostendorf18} adapts to environmental factors, so it could potentially draw on independent experiences with female speakers and with lawyer speech in order to initialize a model of adaptation to a new female lawyer \citep[see also][]{mikolov2012context,kleinschmidt2018structure}. The psycholinguistic plausibility of these models can be tested in future work.

\bibliography{bibliography}
\bibliographystyle{acl_natbib_nourl}

\clearpage

\appendix
\section{Supplementary Materials for A Neural Model of Adaptation in Reading}

\subsection{Language model}

We used the model trained by \citet{gulordavaetal18}. This model was trained on 90 million words of English Wikipedia articles. It had two LSTM layers with 650 hidden units each, 650-dimensional word embeddings, a learning rate of 20, a dropout rate of 0.2 and a batch size 128, and was trained for 40 epochs (with early stopping).

\subsection{Analysis of reading times}
We fit linear mixed-effects models (LMEM) using the \emph{lme4} R package \cite{r-lme4}. Our largest LMEM contained random intercepts for items and subjects, and fixed effects and by-subject random slopes for word length, sentence position, non-adaptive surprisal from the base model, and adaptive surprisal. The full LMEM formula in R notation is as follows:

\begin{quote}
RT $\sim$ word\_length + sentence\_position + non-adaptive\_surprisal + adaptive\_surprisal + (1$\mid$word) + (1 + word\_length + sentence\_position + non-adaptive\_surprisal + adaptive\_surprisal $\mid$subject)
\end{quote}

To assess whether a predictor significantly contributes to the model's fit to the data, we used a likelihood ratio test comparing an LMEM that includes that predictor with an LMEM that does not. 

\subsection{\citet{finejaeger16} simulation}

We reproduce here the examples of the materials from the introduction. The critical region (where surprisal was evaluated to assess adaptation to the higher probability of reduced relative clauses) is underlined.

\ex.\textit{Ambiguous:}\\
The experienced soldiers warned about the dangers \underline{conducted the midnight} raid.

\ex.\textit{Unambiguous:}\\
The experienced soldiers who were warned about the dangers \underline{conducted the midnight} raid.
    
Our replication of their experiment compiled their 80 fillers and 40 critical items into 16 lists (item orders). Four randomized orderings were unique, four orderings had the same items in each position as the first four but with opposite conditions for each critical item, and each of those eight total lists were also presented in reverse order, producing 16 stimulus lists. Each stimulus list contained an ambiguous or an unambiguous version of each critical item but not both, and all filler sentences were present in each list.

\subsection{Dative alternation simulation}
\begin{figure}[t]
    \includegraphics[width=.48\textwidth]{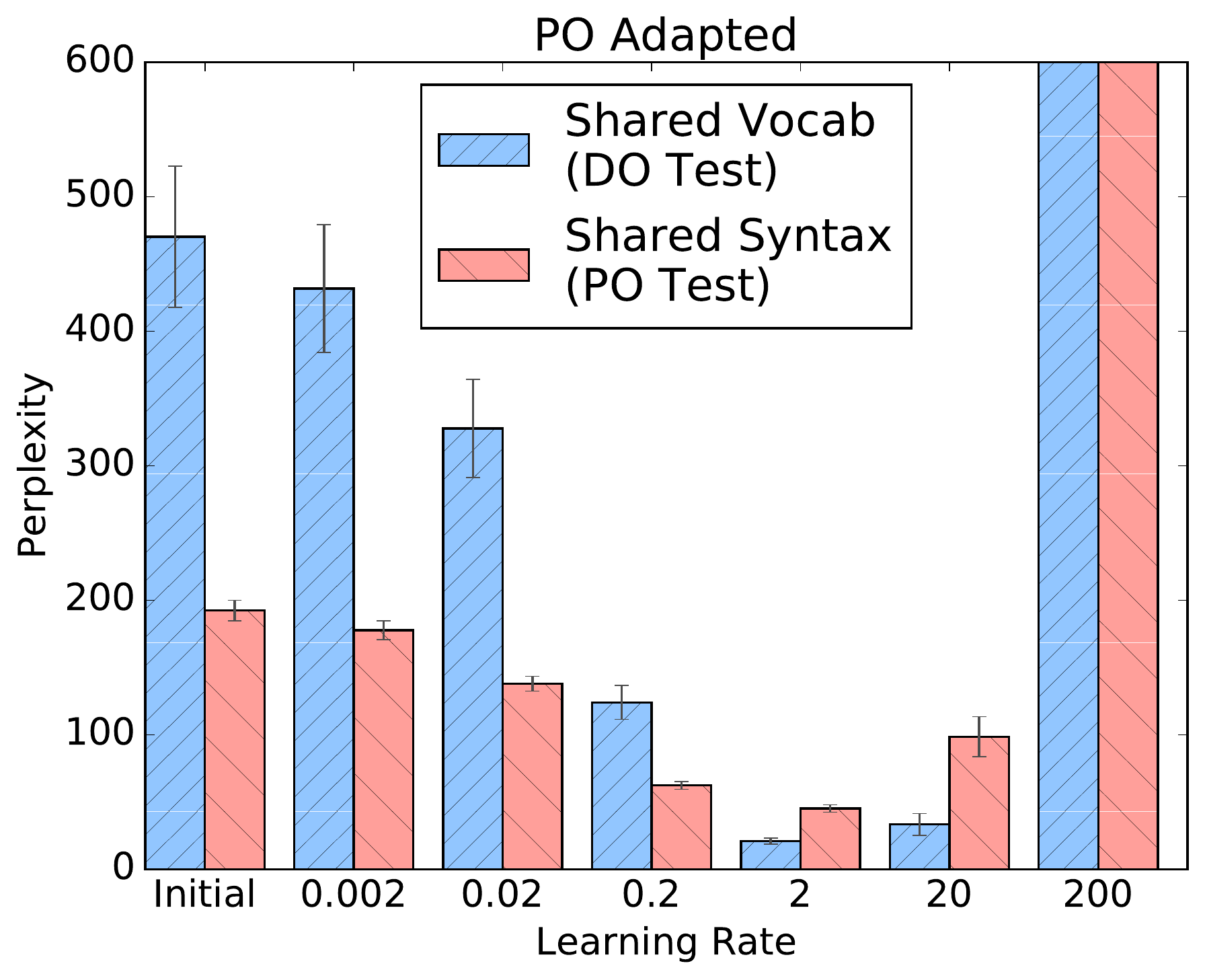}
    \caption{Learning rate influence over syntactic and lexical adaptation. The initial non-adaptive model performance is equivalent to the performance when using a learning rate of 0; the learning rate of 200 resulted in perplexity in the billions.}\label{fig:learnrate-po}
  \end{figure}

We repeated the dative DO adaptation experiment described in the text 10 times, with different critical items and filler sentences in a randomized order for each iteration; we report averaged results and plot the means and standard deviations in our bar charts. We also conducted similar experiments with the roles of the two constructions reversed (the adaptation set included PO instead of DO sentences). The results were very similar to the DO results we report in the paper (see Figure \ref{fig:learnrate-po}). As in the DO adaptation variant, the model initially assigns a lower probability to DO constructions which is the reason behind the non-adaptive model's different performance on the lexical adaptation test set (DO here, PO in the paper) compared with the syntactic adaptation test set (PO here, DO in the paper). In the PO adapted model, at the optimal learning rate, lexical adaptation was sufficient to overcome this syntactic pre-training bias. 

\end{document}